\documentclass[letterpaper, 10 pt, journal, twoside]{IEEEtran}
\usepackage{amsfonts}
\usepackage{times}
\usepackage[numbers,sort&compress]{natbib}

\usepackage{multicol}
\usepackage[bookmarks=true]{hyperref}
\usepackage{colortbl}

\usepackage{mathtools}
\usepackage[utf8]{inputenc}
\usepackage{xcolor}
\usepackage{graphicx}
\usepackage{dirtytalk}
\usepackage[acronyms, shortcuts]{glossaries}
\glsdisablehyper

\usepackage{booktabs}
\usepackage{float}
\usepackage{array}
\usepackage[font=small]{subcaption}
\usepackage{multirow}
\usepackage{snotez}     %
\usepackage[capitalize,noabbrev,nameinlink]{cleveref}
\Crefname{equation}{}{}
\Crefname{algocfline}{Algorithm}{Algorithms}
\Crefname{algocf}{line}{lines}
\Crefname{assumption}{Assumption}{Assumptions}
\crefrangeformat{assumption}{Assumptions~#3#1#4-#5#2#6}
\usepackage{duckuments}
\usepackage{siunitx}
\usepackage{empheq}
\usepackage{cancel}

\usepackage{amsthm}
\usepackage{nccmath}
\usepackage{balance}

\glsdisablehyper

\newcommand{\makiescale}{0.46}
\newcommand{\largermakiescale}{0.47}

\newcommand{\revised}[1]{#1}

\newcommand{\revisedFinal}[1]{#1}

\newcommand{\atrealtime}[2]{{#2}_{#1}}
\newcommand{\realstate}{\hat{\state}}
\newcommand{\realstatetraj}{\hat{\xtraj}}

\newacronym{posg}{POSG}{partially observed stochastic game}
\newacronym{mpgp}{MPGP}{model-predictive game-play}
\newacronym{rl}{RL}{reinforcement learning}
\newacronym{marl}{MARL}{multi-agent reinforcement learning}
\newacronym{pac}{PAC}{probably approximately correct}
\newacronym{mcp}{MCP}{mixed complementarity problem}
\newacronym[plural=gne,firstplural=generalized Nash eqiluibria (GNE)]{gne}{GNE}{generalized Nash equilibrium}
\newacronym{gnep}{GNEP}{generalized Nash equilibrium problem}
\newacronym{mpec}{MPEC}{mathematical program with equilibrium constraints}
\newacronym{kkt}{KKT}{Karush-Kuhn-Tucker}
\newacronym{tro}{T-RO}{IEEE Transactions on Robotics}
\newacronym{ijrr}{IJRR}{International Journal of Robotics Research}
\newacronym{rss}{RSS}{Robotics: Science and Systems}
\newacronym{mle}{MLE}{maximum likelihood estimation}

\makeatletter
\usepackage{xspace}
\DeclareRobustCommand\onedot{\futurelet\@let@token\@onedot}
\def\@onedot{\ifx\@let@token.\else.\null\fi\xspace}
\def\eg{e.g\onedot} 
\def\ie{i.e\onedot} 
\def\cf{cf\onedot}

\makeatother

\newcommand{\jth}{j^\text{th}}

\newcommand{\beliefParam}{\theta}
\newcommand{\branchTime}{t_b}

\newcommand{\share}{c}
\newcommand{\constraint}{h}

\newcommand{\numHypotheses}{K}

\newcommand{\A}{\mathrm{R}}
\newcommand{\B}{\mathrm{H}}

\newcommand{\state}{x}

\newcommand{\control}{u}

\newcommand{\traj}{z}
\newcommand{\btraj}{\boldsymbol{\traj}}

\newcommand{\btrajA}{\btraj^{\A}}
\newcommand{\btrajB}{\btraj^{\B}}
\newcommand{\utraj}{\boldsymbol{u}}

\newcommand{\xtraj}{\boldsymbol{x}}

\newcommand{\cost}{J}

\newcommand{\lagrangian}{\mathcal{L}}
\newcommand{\lmConstraint}{\lambda}
\newcommand{\lmShare}{\rho}
\newcommand{\lowerBounds}{v_\textrm{lo}}
\newcommand{\upperBounds}{v_\textrm{up}}
\newcommand{\mcpVariables}{v}
\newcommand{\mcpFunction}{G}
\newcommand{\solutionset}{\mathcal{S}}

\newcommand{\beliefparam}{\beliefParam}

\newcommand{\horizon}{T}

\newcommand{\ra}{\rightarrow}
\newcommand{\R}{\mathbb{R}}

\newtheoremstyle{mystyle}%
  {}%
  {}%
  {}%
  {}%
  {\bfseries}%
  {.}%
  { }%
  {}%
\theoremstyle{mystyle}

\newtheorem{definition}{Definition}

\newcommand{\identity}{\mathbf{I}}
\newcommand{\entropy}{\mathcal{H}}

\let\originalleft\left
\let\originalright\right
\renewcommand{\left}{\mathopen{}\mathclose\bgroup\originalleft}
\renewcommand{\right}{\aftergroup\egroup\originalright}

\usepackage[font=small]{caption}
\begin{document}

\title{Contingency Games for Multi-Agent Interaction}

\author{
{\fontsize{9.9}{9.9}\selectfont
Lasse Peters\IEEEauthorrefmark{1},
Andrea Bajcsy\IEEEauthorrefmark{2},
Chih-Yuan Chiu\IEEEauthorrefmark{3},
David Fridovich-Keil\IEEEauthorrefmark{4},
Forrest Laine\IEEEauthorrefmark{5},
Laura Ferranti\IEEEauthorrefmark{1},
Javier Alonso-Mora\IEEEauthorrefmark{1}
}
\thanks{Manuscript received: August, 17, 2023; Revised November, 12, 2023; Accepted December, 20, 2023.}%
\thanks{This paper was recommended for publication by Editor G. Venture
upon evaluation of the Associate Editor and Reviewers' comments.
}
\thanks{
This work is funded in part by the European Union (ERC, INTERACT, 101041863). Views and opinions expressed are however those of the author(s) only and do not necessarily reflect those of the European Union or the European Research Council Executive Agency. Neither the European Union nor the granting authority can be held responsible for them.
}
\thanks{
This work received support from the Dutch Science Foundation NWO-TTW within Veni project HARMONIA (18165).
}
\thanks{
This work was supported by the National Science Foundation under Grant No. 2211548.
}
\thanks{
Affiliation: 
\IEEEauthorrefmark{1}Department of Cognitive Robotics, TU Delft, 2628 CD Delft, Netherlands (\{l.peters, l.ferranti, j.alonsomora\}@tudelft.nl).
\IEEEauthorrefmark{2}{Robotics Institute,  Carnegie Mellon University, 5000 Forbes Avenue, Pittsburgh, PA 15213, USA (abajcsy@cmu.edu.).}
\IEEEauthorrefmark{3}{Department of EECS at UC Berkeley, 337 Cory Hall, Berkeley, CA 94720, USA (chihyuan\_chiu@berkeley.edu).}
\IEEEauthorrefmark{4}{Department of Aerospace Engineering
and Engineering Mechanics, UT Austin, Austin,
TX 78712, USA (dfk@utexas.edu).}
\IEEEauthorrefmark{5}{Department of Computer Science, Vanderbilt University, 1400 18th Avenue South, Nashville, TN 37212, USA (forrest.laine@vanderbilt.edu).
}
Corresponding author: Lasse Peters.
}
\thanks{Digital Object Identifier (DOI): see top of this page.}
} %

\markboth{Journal of \LaTeX\ Class Files,~Vol.~14, No.~8, August~2021}%
{Shell \MakeLowercase{\textit{et al.}}: A Sample Article Using IEEEtran.cls for IEEE Journals}

\maketitle
\markboth{Preprint Version}
{Peters \MakeLowercase{\textit{et al.}}: Contingency Games} %

\begin{abstract}
Contingency planning, wherein an agent generates a set of possible plans conditioned on the outcome of an uncertain event, is an increasingly popular way for robots to act under uncertainty. 
In this work we take a game-theoretic perspective on contingency planning, tailored to multi-agent scenarios in which a robot's actions impact the decisions of other agents and vice versa. 
The resulting \emph{contingency game} allows the robot to efficiently interact with other agents by generating strategic motion plans conditioned on multiple possible intents for other actors in the scene.
Contingency games are parameterized via a scalar variable which represents a future time when intent uncertainty will be resolved. 
By estimating this parameter online, we construct a game-theoretic motion planner that adapts to changing beliefs while anticipating future certainty.
We show that existing variants of game-theoretic planning under uncertainty are readily obtained as special cases of contingency games. %
Through a series of simulated autonomous driving scenarios, we demonstrate that \revised{contingency games close the gap between certainty-equivalent games that commit to a single hypothesis and non-contingent multi-hypothesis games that do not account for future uncertainty reduction.}

\vspace{0.5em}
\noindent
Website/Code: \url{https://lasse-peters.net/pub/contingency-games}
\smallskip

\end{abstract}


\section{Introduction}
\label{sec:introduction}

\begin{figure}[tp!]
    \centering
    \includegraphics[width=\linewidth]{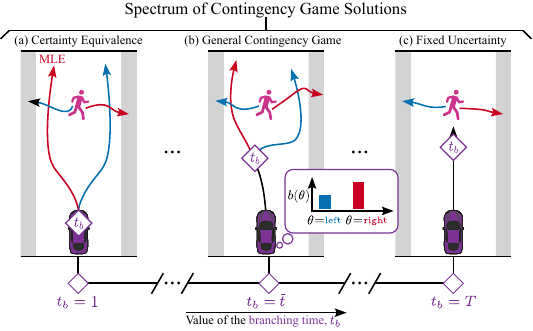}
    \caption{
    A vehicle approaching a jaywalking pedestrian with uncertain intent.
    (a)~A risk-taking driver may gamble for the most likely outcome, ignore uncertainty, and pass on the left.
    (c)~A risk-averse driver may hedge against all outcomes by bringing their vehicle to a full halt, waiting for the situation to resolve.
    (b)~An experienced driver may realize that this uncertainty will resolve in the near future~(at time~$\branchTime$) and thus commit to an \emph{immediate plan} that can be continued safely and efficiently under both outcomes.
    Our contingency games formalize this middle ground between the two extremes.
    }
    \label{fig:front_fig}
\end{figure}

\IEEEPARstart{I}{magine} you are driving and you see a pedestrian in the middle of the road as shown in~\cref{fig:front_fig}.
The pedestrian is likely to continue walking to the right, but you also saw them turning their head around; so maybe they want to walk back to the left?
You think to yourself, \say{If the pedestrian continues to the right, I just need to decelerate slightly and can safely pass on the left; but if they suddenly turn around, I need to brake and pass them on the right.}
Moreover, you understand that your actions influence the pedestrian's decision regarding whether and how quickly to cross the street.
You decide to take your foot off the gas pedal and drive forward, aiming to pass the pedestrian on the left, but you are ready to brake and swerve to the right should the pedestrian turn around.

This example captures three important aspects of real-world multi-agent reasoning:
(\romannumeral 1)~strategic interdependence of agents' actions due to their (partially) conflicting intents---\eg, the pedestrian's actions do not only depend on their own intent but also on your actions, and vice versa;
(\romannumeral 2)~accounting for uncertainty---\eg, how likely is it that the pedestrian wants to move left or right?;
and~(\romannumeral 3)~planning contingencies---\eg, 
by anticipating that uncertainty will be resolved in the future, a driver can commit to an \emph{immediate plan}~(shown in black in \cref{fig:front_fig}b) that can be continued safely and efficiently under each outcome (shown in red and blue in~\cref{fig:front_fig}b).

In this paper, we formalize this kind of reasoning by introducing \textit{contingency games}: a mathematical model for planning contingencies through the lens of dynamic game theory.
Specifically, we focus on \ac{mpgp} settings, wherein an ego agent (\ie, a robot) plans by choosing its future trajectory in strategic \emph{equilibrium} with those of other nearby agents (e.g., humans) at each planning invocation.
Importantly, the ego agent must account for any uncertainty about other agents’ intents---\ie, their optimization objectives---when solving this game.
For computational tractability, most established methods 
take one of two approaches. One class of methods
ignores uncertainty by performing \ac{mle} and solving
a certainty-equivalent game 
\cite{sadigh2016planning,mehr2023maximum,liu2022learning}.
The other class accounts for uncertainty by planning with a full distribution---or ``belief''---conservatively assuming that intent uncertainty will never be resolved during the planning horizon~\cite{laine2021multi,le2021lucidgames}. 
Our \textbf{main contribution} is a game-theoretic interaction model that bridges the gap between these two extremes:
\begin{quote}{A contingency game is a model for \emph{strategic interactions} which allows a robot to consider the full \emph{distribution} of other agents' intents while \emph{anticipating intent certainty} in the near future.}
\end{quote}

Importantly, unlike existing formulations of trajectory games with parametric uncertainty~\cite{le2021lucidgames,laine2021multi,tian2022safety}, contingency games capture the fact that future belief updates will reduce uncertainty and hence, eventually, the \emph{true} intent of the human will be clear at a future ``branching'' time, $t_b$.
As a result, solutions of contingency games are \emph{conditional} robot strategies which take a tree structure as shown in \cref{fig:front_fig}b.
The \say{trunk} of the conditional plan encodes decisions that are made before certainty is reached (before $t_b$). %
After $t_b$, %
the robot generates separate conditional trajectories for each possibility $\beliefParam \in \Theta$.

Beyond our main contribution of an uncertainty-aware game-theoretic interaction model,
we also
(\romannumeral 1) show how general-sum $N$-player contingency games can be transformed into \aclp{mcp}, for which off-the-shelf solvers \cite{dirkse1995path} are available
and (\romannumeral 2) discuss how beliefs and branching times may be estimated online for receding-horizon operation.
We also highlight the desirable modeling flexibility of contingency games as a function of the parameter $\branchTime$, recovering certainty-equivalent games on one extreme, and conservative solutions on the other (see \Cref{fig:front_fig}).
Through a series of simulation experiments, we demonstrate \revised{that contingency games close the gap between these two extremes, and highlight the utility of estimating the branching time online.}

\section{Related Work}
\label{sec:related}

\subsection{Game-Theoretic Motion Planning}
\label{subsec: Game-Theoretic Planning}

Game-theoretic planning has become increasingly popular in interactive robotics domains like autonomous driving \cite{sadigh2016planning, fisac2019hierarchical, so2022mpogames}, drone racing \cite{spica2020real}, and shared control \cite{music2020haptic} due to its ability to model influence among agents.
A crucial axis in which prior works differ is in the modeling of the robot's uncertainty regarding other agents' objectives, dynamics, and state at each time. Methods which assume no uncertainty in the trajectory game model (e.g., taking the most probable hypothesis as truth) result in the simplest game formulations, and have been explored extensively 
\cite{fridovich2020efficient, le2022algames, zhu2022sequential, mehr2023maximum}.
However, in real-world settings it is unrealistic to assume that a robot has full certainty, especially with respect to other agents' intents.
Instead, robots will often maintain a probability distribution, or ``belief,'' over uncertain aspects of the game. 

There are several ways in which such a belief can be incorporated in a trajectory game.
On one hand, the robot could simply optimize for its expected cost under the distribution of uncertain game elements.
We call this as a ``fixed uncertainty'' approach, since the game ignores the fact that as the game evolves, the robot could gain information leading to belief updates \cite{le2021lucidgames,laine2021multi,tian2022safety}.
While these methods do utilize the robot's uncertainty, they often lead to overly conservative plans because the robot cannot reason about future information that would make it more certain (i.e., confident) in its decisions. 

On the other hand, the agents in a game may reason about how their actions could lead to information gain; we refer to this as ``dynamic uncertainty.'' %
Games which exactly model dynamic information gain are inherently more complex, and are generally intractable to solve, especially when the belief space is large and has non-trivial update dynamics~\cite{bernstein2002complexity,goldsmith2007competition}.
Recent methods attempted to alleviate the computational burden of an exact solution via linear-quadratic-Gaussian approximations of the dynamics, objectives, and beliefs \cite{schwarting2021stochastic}.
While the computational benefits of such approximations are significant, they introduce artifacts that make them inappropriate for scenarios such as that shown in \cref{fig:front_fig} in which uncertainty is fundamentally multimodal.
It remains an open challenge to solve ``dynamic uncertainty'' games tractably. 

Our \textit{contingency games} approach presents a middle ground between these paradigms via a belief-update model simple enough to compute exact solutions to the resulting dynamic-uncertainty game, and realistic enough to generate intelligent behavior that anticipates future uncertainty reduction.

\subsection{Non-Game-Theoretic Contingency Planning} \label{subsec: Contingency Planning}
\revised{There is a growing literature of non-game-theoretic interaction planners~\cite{wang2022social}, including various flavors of contingency planning.
Online-optimization-based approaches primarily focus on predict-then-plan contingency planning \cite{hardy2013contingency, zhan2016non, chen2022interactive, nair2022stochastic}. 
Recent learning-based contingency planners leverage deep neural networks to generate human predictions conditioned on candidate robot plans 
\cite{cui2021lookout, tolstaya2021identifying}; 
a robot plan is then selected via methods like sampling-based model-predictive control~\cite{cui2021lookout}, dynamic programming~\cite{chen2023tree}, or neural trajectory decoders~\cite{tolstaya2021identifying}}.

Other approaches draw inspiration from deep reinforcement learning to accomplish both intention prediction and motion planning. 
To predict multi-agent trajectories in the near future, \citet{packer2022anyone} and \citet{rhinehart2019precog} construct a flow-based generative model and a likelihood-based generative model, respectively. 
Meanwhile, \citet{rhinehart2021contingencies} develop an end-to-end contingency planning framework for both intention prediction and motion planning. 
We bring the notion of contingency planning to a different modeling domain---dynamic game theory---to extend its capabilities to handle ``dynamic uncertainty'' in strategic interactions. \revised{This game-theoretic perspective captures interdependent behavior by modeling other agents as minimizing their own cost as a function of the decisions of all players in the scene.}

\section{Formalizing Contingency Games}
\label{sec:problem}
In this paper, we consider settings where a game-theoretic interaction model is not fully specified.
For example, an autonomous car may not be sure if a nearby pedestrian intends to jaywalk; an assistive robot may not know which tool a surgeon will wish to use next.
In such instances, the robot (agent $\A$) can construct a dynamic game in which components of the model depend upon an unknown parameter $\beliefParam \in \Theta$, \revised{with $|\Theta| = \numHypotheses < \infty$}.
When the robot has some prior information---\eg, from observations of past behavior---it can maintain a probability distribution or \emph{belief}, $b(\beliefParam)$, over the set of possible games.
Naturally, however, this belief changes as a function of the human's behavior \emph{and} the robots actions.
It is this \emph{dynamic} nature of the uncertainty that the robot can exploit to come up with more efficient plans.
In the formal description below, we adopt a two-player convention for clarity. 
\revised{We note, however, that the formalism can incorporate more players as illustrated in \cref{sec:driving-scenarios}.}

\smallskip
\noindent \textbf{Approximations and modeling assumptions.}
Contingency games approximate a game which exactly models dynamic uncertainty, which are generally intractable to solve.
Specifically, we introduce the following key modeling assumptions to facilitate tractable online computation:
\begin{enumerate}
    \item We assume unilateral uncertainty with discrete support, \revised{i.e. $|\Theta| = \numHypotheses$}. That is, while the robot $\A$ has uncertainty in the form of a discrete probability mass function $b(\beliefParam)$, the human $\B$ acts rationally under the true hypothesis $\hat{\beliefParam}$.
    \item We assume the robot has access to (an estimate of) the so-called branching time, $t_b$, which models the future time at which additional state observations will have resolved all uncertainty about the initially unknown~$\theta$.
    \item We simplify the robot's belief dynamics. Instead of capturing the exact belief dynamics across the planning horizon, the robot distinguishes two phases: before $t_b$ the belief is fixed at $b(\cdot)$; after $t_b$ the belief collapses to certainty about a single hypothesis.
\end{enumerate}
We envision contingency games to be employed in a receding horizon (\ac{mpgp}) fashion.
In that context, beliefs are updated between each planner invocation and the branching time may vary and can be estimated online.\footnote{
\revised{Note that, despite the use of assumption 3 within our game formulation, we will employ a belief updater that still captures more accurate belief dynamics as we shall discuss in \cref{sec:online-contingency-setup}.}}
We discuss considerations and results for this case in \cref{sec:online-contingency-setup,sec:simulation-results}.

\smallskip
\noindent\textbf{Notation conventions.}
We consider interaction of agents $i \in \{\A, \B\}$ over $\horizon < \infty$ time steps.
In contrast to existing game-theoretic formulations \cite{basar1998gametheorybook,le2022algames,fridovich2020efficient,laine2023computation}, in a contingency game we endow each player with multiple trajectories; one for each hypothesis~$\beliefparam$.
At each $t \in [\horizon] = \{1, 2, \dots, \horizon\}$ and hypothesis~$\beliefParam\in\Theta$ the \emph{state} of the game is comprised of separate variables for each agent~$i$, \ie, $\state_{\beliefParam, t} := (\state^\A_{\beliefparam, t}, \state^\B_{\beliefparam, t})$.
Each agent~$i$ begins at a fixed initial state $\hat{\state} := (\hat{\state}^\A, \hat{\state}^\B)$, \ie, $\state^i_{\revised{\beliefParam, 1}} = \hat \state^i$, which evolves over time as a function of that agent's control action, $\control^i_{\revised{\beliefparam, t}}$.
States and control actions are assumed to be real vectors of arbitrary, finite dimension.
For brevity, we introduce the following shorthand: we use $\traj_{\theta,t}^i := (\state_{\theta, t}^i, \control_{\theta, t}^i)$ to denote a state-control tuple, we use boldface to denote aggregation over time, \eg $\btraj_\beliefparam^i := (\traj_\beliefparam^i)_{t \in [\horizon]}$, we omit player indices to denote aggregation, \eg $\btraj_\beliefParam = (\btrajA_\beliefParam, \btrajB_\beliefParam)$, and we denote the finite collection of all $\numHypotheses = |\Theta|$ trajectories for player~$i$ as $\btraj^i_\Theta = (\xtraj^i_{\Theta}, \utraj^i_{\Theta}) := (\btraj^i_\beliefParam)_{\beliefParam\in\Theta}$.

\smallskip
\noindent\textbf{Contingency game formulation.}
With these conventions in place, we formulate a contingency game as follows.
The robot wishes to optimize its expected performance over all hypotheses~\cref{eqn:Aobj} while restricting all contingency plans to be feasible with respect to hypothesis-dependent constraints~$\constraint_\beliefparam$~\cref{eqn:Aconstraint} and \revised{enforcing the \emph{contingency constraint}~\cref{eqn:Ashare} that the first $\branchTime-1$ control inputs must be identical across all hypotheses~$\beliefparam \in \Theta$}:
\begin{subequations}
\label{eqn:Aoptproblem}
\begin{alignat}{3}
    \label{eqn:Aobj}
    &\A: \hspace{2mm}  &\solutionset^\A\left(\btrajB_\Theta\right) :=&\arg\min_{\btrajA_{\Theta}}  &&\quad \sum_{\beliefParam \in \Theta} b(\beliefParam) \cost^\A(\btrajA_{\beliefParam}, \btrajB_{\beliefParam}) \\ 
    \label{eqn:Aconstraint}
    & && \hspace{8mm}  &&\constraint_\beliefParam^\A(\btrajA_\beliefParam, \btrajB_\beliefParam) \geq 0,~\forall \beliefParam \in \Theta\\
    \label{eqn:Ashare}
    & && \hspace{8mm} &&\share(\utraj^\A_{\Theta}; \branchTime) = 0.
\end{alignat}
\end{subequations}

Simultaneously, the robot interacts with $\numHypotheses$ versions of agent $\B$, each of which is guided by a different intent~$\beliefParam$ which parameterizes both the hypothesis-dependent cost~$\cost^\B_\beliefParam$~\cref{eqn:Bobj} and constraints~$\constraint^\B_\beliefparam$~\cref{eqn:Bconstraint}:
\begin{subequations}\label{eqn:Boptproblem}
\begin{empheq}[left=\forall \beliefParam \in \Theta: \empheqlbrace]{align}
    \label{eqn:Bobj}
    \B_{\beliefParam}: \solutionset^{\B_\beliefparam}\left(\btrajA_\beliefparam\right):=\arg\min_{\btrajB_{\beliefParam}}\hspace{1pt}
    &\cost_\beliefParam^\B(\btrajA_\beliefParam, \btrajB_\beliefParam)\\
    &\constraint_\beliefParam^\B(\btrajA_\beliefParam, \btrajB_\beliefParam) \geq 0.
    \label{eqn:Bconstraint}
\end{empheq}
\end{subequations}
In this contingency game formulation, it is important to appreciate that the robot's strategy depends on the distribution $b(\beliefParam)$ and \emph{all} the trajectories~$\btraj_\Theta = (\btraj_\beliefParam)_{\beliefParam \in \Theta}$, while $\B_\beliefParam$'s strategy depends \emph{only} on the trajectories under hypothesis~$\beliefParam$,~$\btraj_\beliefParam$.
Since the objectives of $\A$ and $\B$ may in general conflict with one another, and we model agents as being self-interested, such a game is termed \emph{noncooperative}.

Taken together, optimization problems~\cref{eqn:Aoptproblem,eqn:Boptproblem} take the form of a~\ac{gnep}.
\say{Solutions} to this problem are defined as follows.

\begin{definition}
\label{def:gne}
(Generalized Nash Equilibrium, \cite[Ch. 1]{facchinei2003finite}).
A trajectory profile~$(\btraj_\Theta^{\A*}, \btraj_\Theta^{\B*})$ is a \emph{\ac{gne}} of the game from \cref{eqn:Aoptproblem,eqn:Boptproblem} if and only if
\begin{align*}
    \btraj^{\A*}_\Theta \in \solutionset^\revised{\A}\left(\btraj_\Theta^{\B*}\right)\quad &\text{and}\quad \bigwedge_{\beliefParam\in\Theta} \btraj^{\B*}_\beliefparam \in \solutionset^{\revised{\B_\beliefparam}}\left(\btraj^{\A*}_\theta\right).
\end{align*}
\end{definition}
In practice, we relax \cref{def:gne} and only require \emph{local} minimizers of~\cref{eqn:Aoptproblem,eqn:Boptproblem}.
Under appropriate technical qualifications, such local equilibria are characterized by first and second order conditions commensurate with concepts in optimization.
Readers are directed to \cite{facchinei2003finite} for further details.

A solution method for contingency games is discussed in~\cref{sec:approach}; but first, we take a step back to build intuition about the behavior induced by the proposed interaction model.

\section{Features of Contingency Game Solutions}
\label{sec:contingency-features}

\begin{figure*}[t!]
\centering
\includegraphics[width=\textwidth]{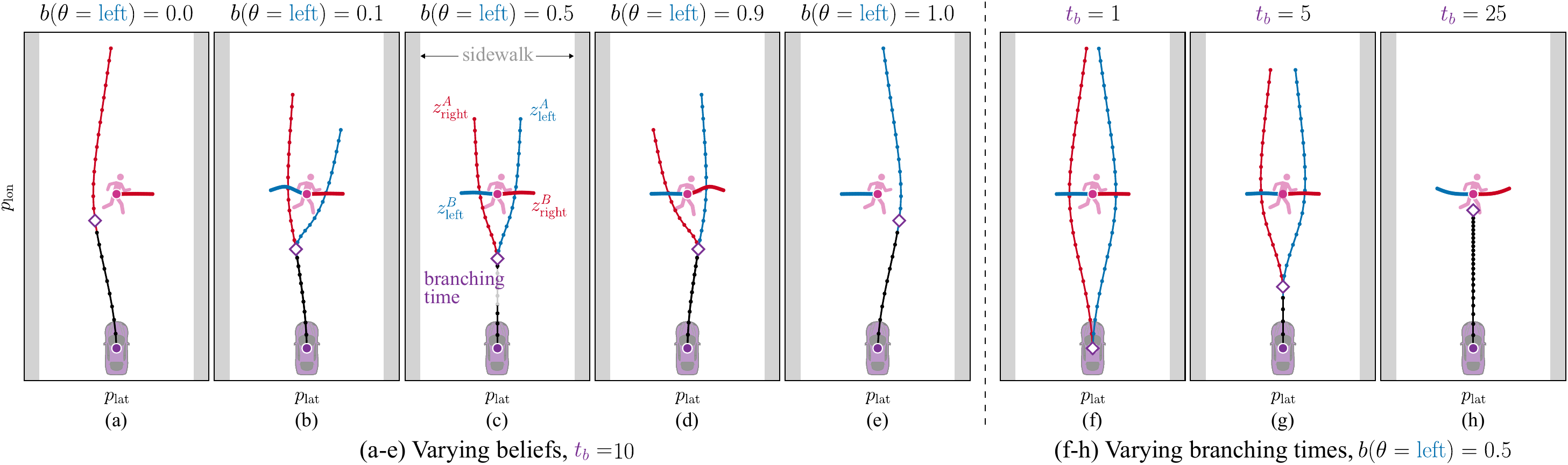}
\caption{\revised{Contingency game solutions for the jaywalking scenario at varying beliefs and branching times.}}
\label{fig:strategy-matrix}
\end{figure*}

\smallskip
\noindent \textbf{Scenario 1: driving near a jaywalking pedestrian.}
Consider %
the scenario shown in \cref{fig:strategy-matrix}.
Here, the robot ($\A$) wants to move forward but does not know if the pedestrian~($\B$) wants to reach the left or right side of the road.
To model this ambiguity, let~$\beliefParam \in \Theta := \{{\mathrm{left}}, {\mathrm{right}}\}$.
Robot $\A$ plans a trajectory for each hypothesis, \ie~$\btrajA_{\Theta} := (\btrajA_{{\mathrm{left}}}, \btrajA_{{\mathrm{right}}})$, each of which is tailored to react appropriately to human $\B_\beliefParam$ in the corresponding scenario. 
Similarly, $\B$ maintains~$\btrajB_{\Theta} := (\btrajB_{{\mathrm{left}}}, \btrajB_{{\mathrm{right}}})$, which capture its ``ground truth'' behavior under each hypothesis. Each of $\B$'s trajectories will react to the corresponding trajectory of $\A$.
We model the robot as a kinematic unicycle and the pedestrian as a planar point mass.
To ensure collision avoidance, the robot must pass behind the pedestrian, \ie~not between the pedestrian and its (initially unknown) goal position.
We capture all of these constraints via a single vector-valued function~$\constraint_\beliefParam^i(\btrajA_\beliefParam, \btrajB_\beliefParam) \geq 0$, which explicitly depends on hypothesis $\beliefParam$.

How each agent acts in a contingency game formulation of this problem is largely affected by two quantities: (\romannumeral1) the initial belief that the robot holds over the human's intent and (\romannumeral2) the branching time which models the time at which the robot will get certainty.
We discuss the role of both quantities below.

\smallskip
\noindent \textbf{Qualitative behavior: Role of belief.}
First, we keep the branching time fixed and analyze the contingency plans for a suite of beliefs.
\mbox{Figures~\ref{fig:strategy-matrix}a-e} show how both the robot's contingency plan and the pedestrian's reaction change as a function of the robot's intent uncertainty. 
At extreme values of the belief, the robot is certain which hypothesis is accurate and the contingency strategy is equivalent to that of a single-hypothesis game with intent certainty.
At intermediate beliefs, the contingency game balances hypotheses' cost according to their likelihood, yielding interesting behaviors: in \cref{fig:strategy-matrix}d the robot plans to inch forward at first (black), but biases its initial motion towards the scenario where the pedestrian will go left (blue). Nevertheless, it still generates a plan for the less likely event that the pedestrian goes right (red).%

\smallskip
\noindent \textbf{Qualitative behavior: Role of branching time.}
Next, we assume that the robot's belief is always a uniform distribution, and vary the contingency game's $\branchTime$ parameter. 
Figures~\ref{fig:strategy-matrix}f-h show how extreme values of this parameter automatically recover existing variants of game-theoretic planning under uncertainty: certainty-equivalent games 
\cite{sadigh2016planning, mehr2023maximum, liu2022learning}
and non-contingent games that plan in expectation \cite{laine2021multi, tian2022safety}.
At one extreme, when~\revised{$\branchTime = 1$}, the contingency constraint~\cref{eqn:Ashare} is removed entirely and we obtain the solutions of the fully observed game under each hypothesis~(see \cref{fig:strategy-matrix}f).
These are precisely the solutions found for the certainty-equivalent games in Figures~\ref{fig:strategy-matrix}a,e.
Note, however, that in this special case the contingency plan 
is not immediately actionable since it first requires the ego agent to commit to a single branch, e.g., by considering only the most likely hypothesis 
\cite{sadigh2016planning,mehr2023maximum,liu2022learning}.
While easy to implement, such an approach can be overly optimistic and can lead to unsafe behavior since the control sequence from the selected branch may be infeasible for another intent hypothesis.
For example, if the robot were to commit to $\beliefParam = \text{left}$ in \cref{fig:strategy-matrix}f, it would be on a collision course with 50\% probability.

By contrast, at the upper extreme of the branching time, $\branchTime = \horizon$, the contingency plan no longer branches and it consists of a single control sequence for the entire planning horizon, \cf~\cite{laine2021multi,tian2022safety}.
As a result, this plan is substantially more conservative since it must trade off the cost under \emph{all} hypotheses while respecting the constraints for any hypothesis with non-zero probability mass: in~\cref{fig:strategy-matrix}g, the ego agent plans to slow down aggressively since its uncertainty about the pedestrian's intent renders both sides of the roadway blocked.

Overall, this analysis shows that contingency games~%
(\romannumeral 1)~unify various popular approaches for game-theoretic planning under uncertainty, and~(\romannumeral 2)~go beyond these existing formulations towards games that consider the \textit{distribution} of other players' intents while \textit{anticipating intent certainty} in the future. 

\smallskip
\noindent \textbf{Quantitative impact of the branching time.}
\begin{figure}
\centering
\includegraphics[scale=\makiescale]{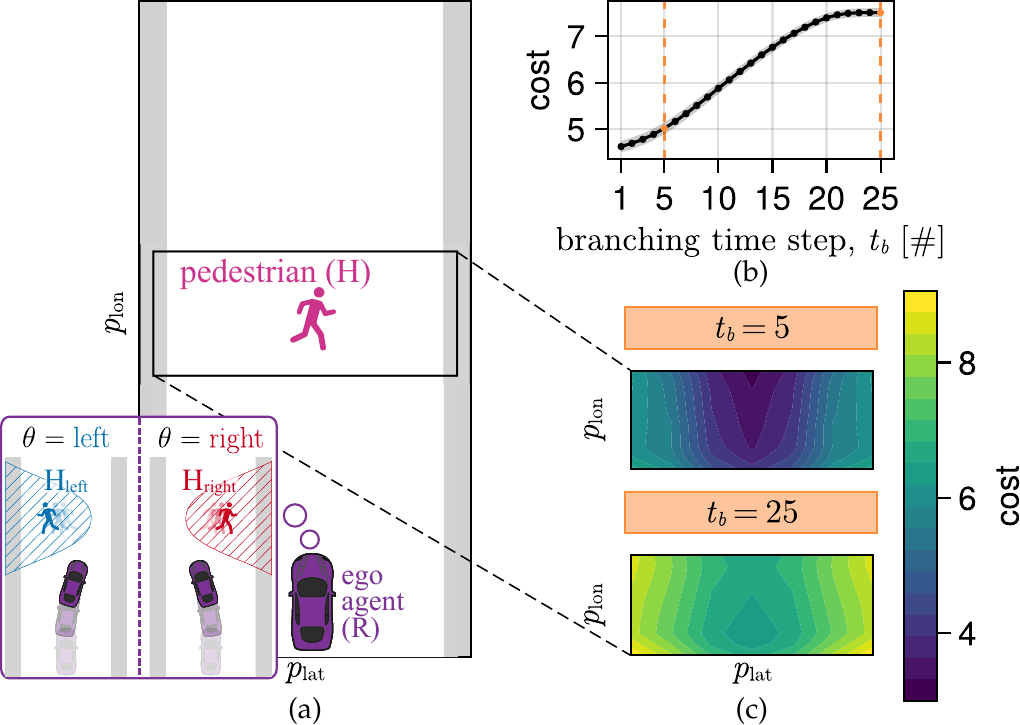}
\caption{Cost of contingency plans in the jaywalking scenario.
(a) state sampling region, (b) cost per $\branchTime$ averaged over all initial states, (c) spatial cost distribution for two fixed $\branchTime$.}
\label{fig:open-loop-cost-gap-spatial}
\end{figure}
Given the central role of the branching time in our interaction model, we further analyze the quantitative impact of this parameter on the robot's contingency plan.
For this purpose, we generate contingency plans across varying branching times, $\branchTime\in\revised{\{1, \ldots, 25\}}$, for each of 70 different initial pedestrian positions sampled from a uniform grid around the nominal position as shown in \cref{fig:open-loop-cost-gap-spatial}a.
\Cref{fig:open-loop-cost-gap-spatial}b shows that, by anticipating future certainty at earlier branching times ($\branchTime = 5$), the robot discovers lower-cost plans than a method that assumes uncertainty will never resolve ($\branchTime = 25$).
Furthermore, the spatial distribution of the cost in \cref{fig:open-loop-cost-gap-spatial}c reveals that robot generates particularly low-cost plans for $\branchTime=5$ if the human is initially in the center of the road.
Here, the contingency plan exploits the fact that---irrespective of the human's true intent---the road will be cleared by the time the robot arrives, \cf~\cref{fig:strategy-matrix}g.
Of course, this analysis pertains to the \emph{open-loop} plan.
However, as we shall demonstrate in \cref{sec:simulation-results}, a performance advantage persists under the added effect of receding-horizon planning.

\section{Transforming Contingency Games into\\Mixed Complementarity Problems}
\label{sec:approach}
Next, we discuss how to compute strategies from this interaction model.
Rather than developing a specialized solver, we demonstrate how contingency games can be transformed into \acp{mcp} for which large-scale off-the-shelf solvers are readily available~\cite{dirkse1995path}.
Our implementation of a game-theoretic contingency planner internally synthesizes such \acp{mcp} from user-provided descriptions of dynamics, costs, constraints, and beliefs.

We begin by deriving the \ac{kkt} conditions for the contingency game in \cref{eqn:Aoptproblem,eqn:Boptproblem}.
Under an appropriate constraint qualification (e.g., Abadie, or linear independence \cite[Ch. 1]{facchinei2003finite}, \cite[Ch. 12]{nocedal2006optimizationbook}), these first-order conditions are jointly necessary for any generalized Nash equilibrium of the game.
The Lagrangians of both players are:
{\fontsize{10}{10}\selectfont
\begin{align}
    \A: \hspace{1mm} 
    &\lagrangian^\A(\btrajA_\Theta, \btrajB_\Theta, \lmConstraint^\A_\Theta, \lmShare) = \lmShare^\top\share(\utraj^\A_{\Theta}; \branchTime)~+\nonumber\\
    & \hspace{5mm} \sum_{\theta \in \Theta}  \Big(b(\beliefParam) \cost^\A(\btrajA_{\beliefParam}, \btrajB_{\beliefParam})
    - {\lmConstraint_\beliefParam^{\A\top}} \constraint_\beliefParam^\A(\btrajA_\beliefParam, \btrajB_\beliefParam)\Big),\\ 
    \B_\beliefParam: \hspace{1mm}
    &\lagrangian^\B_\beliefParam(\btrajA_\beliefParam, \btrajB_\beliefParam, \lmConstraint^\B_\beliefParam)
    = \cost_\beliefParam^\B(\btrajA_\beliefParam, \btrajB_\beliefParam)
    - {\lmConstraint_\beliefParam^{\B\top}} \constraint_\beliefParam^\B(\btrajA_\beliefParam,\nonumber
    \btrajB_\beliefParam),
\end{align}
}
where $\lmShare$ is the Lagrange multiplier for player~$\A$'s contingency constraint, and $\lmConstraint^i_\beliefParam$ are Lagrange multipliers for all other constraints of player~$i$ at hypothesis~$\beliefParam$. 
Denoting complementarity by \say{$\perp$}, we derive the following \ac{kkt} system for all players:
{\fontsize{10}{10}\selectfont
\begin{subequations} \label{eqn:KKTconditions}
\begin{empheq}[left=\forall \beliefParam \in \Theta: \hspace{5mm} \empheqlbrace]{align}
    \nabla_{\btrajA_\beliefParam} \lagrangian^\A(\btrajA_\Theta, \btrajB_\Theta, \lmConstraint^\A_\Theta, \lmShare) &= 0, \\
    \nabla_{\btrajB_\beliefParam} \lagrangian^\B_\beliefParam(\btrajA_\beliefParam, \btrajB_\beliefParam, \lmConstraint^\B_\beliefParam) &= 0, \\
    0 \leq \constraint_\beliefParam^\A(\btrajA_\beliefParam, \btrajB_\beliefParam) \perp \lmConstraint_\beliefParam^\A &\geq 0, \\
    0 \leq \constraint_\beliefParam^\B(\btrajA_\beliefParam, \btrajB_\beliefParam) \perp \lmConstraint_\beliefParam^\B &\geq 0,
\end{empheq}
\vspace{-5pt}
\begin{equation}
\hspace{42mm} \share(\utraj^\A_{\Theta}; \branchTime) = 0.
\end{equation}
\end{subequations}
}
Collectively, \eqref{eqn:KKTconditions} forms an \ac{mcp}, as defined below
\cite[Ch.~1]{facchinei2003finite}.

\begin{definition}[\textbf{Mixed Complementarity Problem}] \label{Def: MCP}
A \acf{mcp} takes the following form: Given $\mcpFunction: \R^d \ra \R^d$, lower bounds $\lowerBounds \in [-\infty, \infty)^d$ and upper bounds $\upperBounds \in (-\infty, \infty]^d$, solve for~$\mcpVariables^\star \in \R^d$ such that, for each~$j \in \{1, \cdots, d\}$, one of the equations below holds:
\begin{align*}
    [\mcpVariables]_j^\star = [\lowerBounds]_j, [\mcpFunction]_j(\mcpVariables^\star) &\geq 0, \\
    [\lowerBounds]_j < [\mcpVariables]_j^\star < [\upperBounds]_j, [\mcpFunction]_j(\mcpVariables^\star) &= 0, \\
    [\mcpVariables]_j^\star = [\upperBounds]_j, [\mcpFunction]_j(\mcpVariables^\star) &\leq 0,
\end{align*}
where~$[\mcpFunction]_j$ denotes the $\jth$ component of $\mcpFunction$, and $[\lowerBounds]_j$ and $[\upperBounds]_j$ denote the $\jth$ components of $\lowerBounds$ and $\upperBounds$, respectively.
\end{definition}

Observe that the~\ac{kkt} conditions~\eqref{eqn:KKTconditions} encode an~\ac{mcp} with variable~$\mcpVariables$, function~$\mcpFunction$, and bounds~$\lowerBounds$,~$\upperBounds$ block-wise defined~(by slight abuse of notation): one block for each~$\beliefParam \in \Theta$,
{\fontsize{10}{10}\selectfont
\begin{subequations}
\label{eq:game-mcp-block-theta}
\begin{align}
    [\mcpVariables]_\beliefParam &= (\btrajA_\beliefParam, \btrajB_\beliefParam, \lmConstraint^\A_\beliefParam, \lmConstraint^\B_\beliefParam), \\
    [\lowerBounds]_\beliefParam &= (-\infty, -\infty, 0, 0), \\
    [\upperBounds]_\beliefParam &= (\infty, \infty, \infty, \infty), \\
    [\mcpFunction(\mcpVariables)]_\beliefParam &= \begin{bmatrix}
    \nabla_{\btrajA_\beliefParam} \lagrangian^\A(\btrajA_\Theta, \btrajB_\Theta, \lmConstraint^\A_\Theta, \lmShare) \\
    \nabla_{\btrajB_\beliefParam} \lagrangian^\B(\btrajA_\beliefParam, \btrajB_\beliefParam, \lmConstraint^\B_\beliefParam) \\
    \constraint_\beliefParam^\A(\btrajA_\beliefParam, \btrajB_\beliefParam) \\
    \constraint_\beliefParam^\B(\btrajA_\beliefParam, \btrajB_\beliefParam)
    \end{bmatrix},
    \label{eq:game-mcp-block-theta-function}
\end{align}
\end{subequations}
}
and an additional block for the contingency constraint
\begin{align}
    \label{eq:game-mcp-block-share}
    [\mcpVariables]_\share \hspace{-.25mm}=\lmShare,
    [\lowerBounds]_\share\hspace{-.25mm}= -\infty,
    [\upperBounds]_\share\hspace{-.25mm}= \infty,
    [\mcpFunction(\mcpVariables)]_\share\hspace{-.25mm}= \share(\utraj^\A_{\Theta}; \branchTime).
\end{align}
To establish that the \ac{mcp} solution is indeed a local equilibrium of the contingency game, sufficient second-order conditions \cite[Thm. 12.6]{nocedal2006optimizationbook} can be checked for \cref{eqn:Aoptproblem} and \cref{eqn:Boptproblem}.

\section{Online Planning with Contingency Games}\label{sec:online-contingency-setup}
We envision contingency games to be deployed in a \ac{mpgp} framework where beliefs and branching times are estimated online.
Next, we discuss considerations for this setting.%

\subsection{Belief Updates}
\label{sec:belief-updates}
\begin{figure}[t!]
    \centering
    \includegraphics[width=0.7\linewidth]{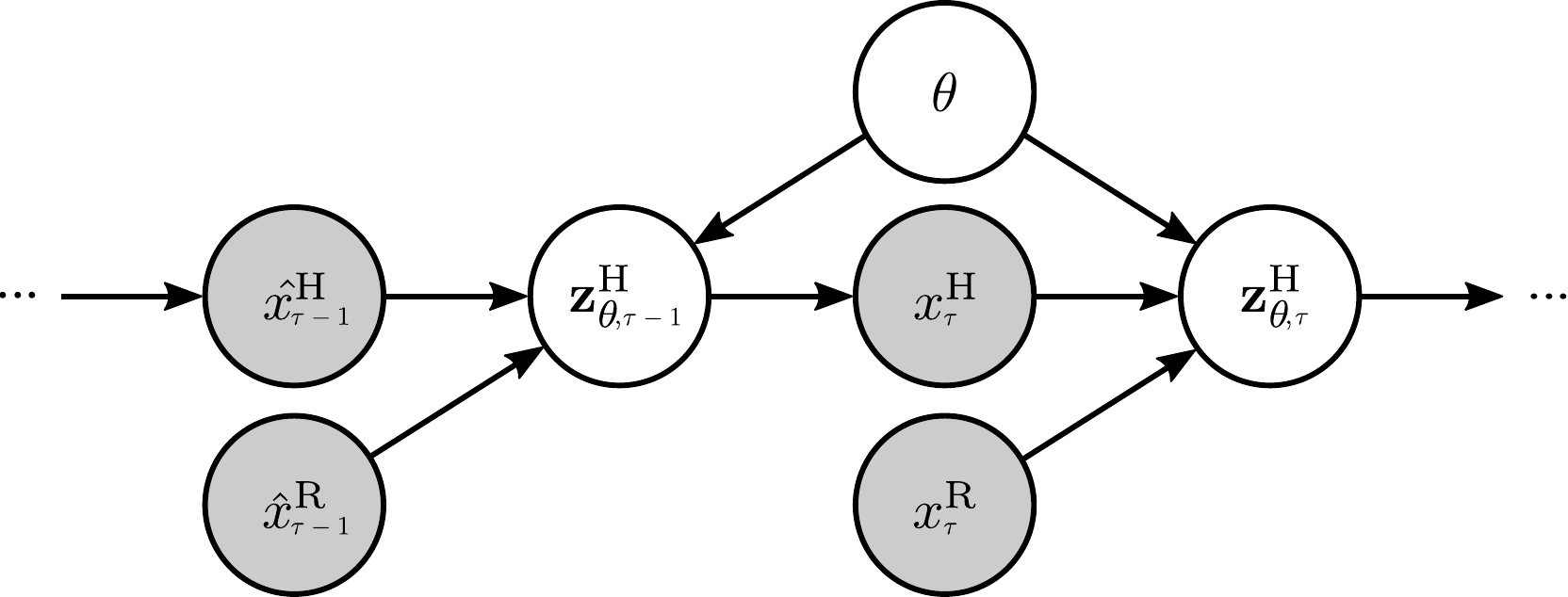}
    \caption{
    Bayesian network modeling the robot's intent inference problem. Shaded nodes represent observed variables.
    }
    \label{fig:ddn}
\end{figure}

While the true human intent $\hat{\beliefParam}$ is hidden from the robot, the robot can utilize observations of past human decisions to update its current belief~$\atrealtime{\tau}{b}(\beliefParam) = P(\beliefParam \mid \realstatetraj)$ about this quantity, where $\realstatetraj := \{(\atrealtime{1}{\realstate^\A}, \atrealtime{1}{\realstate^\B}), \hdots (\atrealtime{\tau}{\realstate^\A}, \atrealtime{\tau}{\realstate^\B})\}$ is the sequence of joint human-robot states observed up until the current time $\tau$. 
We use the model in \cref{fig:ddn} to cast this inference problem in a Bayesian framework. 
In this model, we use our game-theoretic planner to compute jointly a \textit{nominal} state-action trajectory for each agent and for each hypothesis. %
The robot's portion of this solution computed at time $\tau$, ${\btrajA_{\Theta,\tau}}$, serves as our receding-horizon motion plan, and the human's portion of the plan, ${\btrajB_{\beliefParam,\tau}}$, constitutes a nominal receding-horizon prediction of their future actions under each hypothesis~$\theta\in\Theta$.
However, due to bounded rationality \cite{rubinstein1998modeling}, the human may not execute exactly this plan.
Hence, at the next time step $\tau +1$ when we observe a new human physical state, $\atrealtime{\tau+1}{\realstate^\B}$, we treat it as a random emission of the \textit{previous} nominal predictions.
\revised{Similar to prior works~\cite{tolstaya2021identifying,nair2022stochastic}}, in our experiments we assume that human states are distributed according to a Gaussian mixture model with one mode for each hypothesis $\theta$; \ie,~$p(\atrealtime{\tau+1}{\realstate^\B} \mid {\btrajB_{\beliefParam, \tau}}) = \mathcal{N}(\mu_{\theta}, \Sigma)$ where the mean $\mu_\theta$ is the expected human state extracted from the \emph{previous} human prediction, $\btrajB_{\theta,\tau}$,  and~$\Sigma$ is a covariance parameter characterizing human rationality.
In summary, the robot can recursively update their belief via
{\fontsize{10}{10}\selectfont
\begin{align}\label{eq:belief-update}
    \atrealtime{\tau+1}{b}(\beliefParam)
    &= \frac{p(\atrealtime{\tau+1}{\realstate^\B} | {\btrajB_{\beliefParam,\tau}})\atrealtime{\tau}{b}(\theta)}{
        \sum_{\beliefParam' \in \Theta} p(\atrealtime{\tau+1}{\realstate^\B} \mid {\btrajB_{\beliefParam', \tau}}) \atrealtime{\tau}{b}(\beliefParam')
    }.
\end{align}
}

\subsection{Estimating Branching Times}
\label{sec:estimating-tb}
Prior work on non-game-theoretic contingency planning either chooses the branching time as informed by a careful heuristic design~\cite{hardy2013contingency} or through offline analysis of the belief updater~\cite{bajcsy2021analyzing}.
A thorough theoretical analysis on how to choose the branching time~$\branchTime$ in the more general game-theoretic case is beyond the scope of this work.
Nonetheless, to demonstrate the utility of anticipating future intent certainty, we propose a branching time heuristic which only requires access to the previous game solution and current belief.

\smallskip\noindent\textbf{Heuristic branching time estimator.}
Intuitively, the branching time should be lower when the future human actions are more ``distinct'' under each hypothesis.
Our heuristic captures this relationship as follows.
Let $\entropy[\atrealtime{\tau}{b}]~= -\sum_{\beliefParam \in \Theta} \atrealtime{\tau}{b}(\beliefParam)\log_{|\Theta|}(\atrealtime{\tau}{b}(\beliefParam))$ denote the entropy of belief~$\atrealtime{\tau}{b}$.
\revised{Furthermore, let $B(\btrajB_{\Theta,\tau-1}, \beliefParam, k)[\cdot]$ denote the operator that, for a \emph{given} $\beliefParam$, takes the first~$k$ states from the previously-computed  $\btrajB_{\Theta,\tau-1}$ as hypothetical observations and returns the thus updated belief.}
We approximate the branching time as
{\fontsize{8.6}{8.6}\selectfont
\begin{align}\label{eq:heuristic-tb-estimator}
    {t_b}(b_\tau, \btrajB_{\Theta,\tau-1}) = \max_{\beliefParam \in \Theta}&\min_{\revised{k \in \{2,\ldots, T\}}}&& k\\
    &\quad\text{s.t.}&& \entropy[B({\btrajB_{\beliefParam, \tau-1}}, \beliefParam, k)[\atrealtime{\tau}{b}]] \leq \epsilon.\nonumber
\end{align}
}
\revised{Note that the minimum branching time chosen by this heuristic is $\branchTime = 2$, ensuring that the robot has a unique first input to apply during receding-horizon operation.}
Procedurally, this heuristic is straightforward to implement: \revised{for \emph{each} hypothesis, we predict the belief via a hypothetical observation sequence} as if the human (\romannumeral 1) is perfectly rational and (\romannumeral 2) does not re-plan in the future; then, we return the first time at which \revised{\emph{all}} predicted beliefs reach entropy threshold~$\epsilon$\footnote{\revised{A low threshold results in more conservative behavior. As informed by a parameter sweep over $\epsilon$, we choose $\epsilon = 2^{-2}$ for all experiments for best performance.}}.
Assumptions (\romannumeral 1) and (\romannumeral 2) make this heuristic cheap to evaluate since they avoid the need to re-compute game solutions within~\cref{eq:heuristic-tb-estimator}.
While, these approximations may affect the accuracy of the estimator %
we shall demonstrate the utility of this approach in \cref{sec:simulation-results}.

\section{Experimental Setup}\label{sec:experimental-setup}
We wish to study the value of \emph{anticipating} future certainty in game-theoretic planning.
Therefore, we compare our method against a non-contingent game-theoretic baselines on two simulated interaction scenarios in which dynamic uncertainty naturally occurs.

\subsection{Compared Methods}
Beyond the contingency game that uses our branching time heuristic, we consider the following methods, \revised{all of which operate in receding-horizon fashion with online belief updates according to~\cref{eq:belief-update}}.
\revised{Note that, following the discussion in \cref{sec:contingency-features}, all game-theoretic methods below can be understood as a contingency game with a special branching time choice}.

\smallskip\noindent
\revised{
\textbf{Baseline 1: Certainty-equivalent ($t_b = 1$).}
This baseline assumes certainty, making a point estimate by considering only the most probable hypothesis at each time step.
}

\smallskip\noindent
\textbf{\revised{Baseline 2: Fixed uncertainty ($t_b = T$).}}
\revised{Similar to~\cite{laine2021multi}, this baseline ignores future information gains, assuming fixed uncertainty along the entire planning horizon.}

\smallskip\noindent
\textbf{\revised{Baseline 3: MPC}.}
\revised{This baseline uses non-game-theoretic model-predictive control (MPC), forecasting opponent trajectories assuming constant ground-truth velocity.}

\smallskip\noindent
\revised{
\textbf{Contingency game with $\branchTime = 2$.}
To test the utility of our branching time heuristic, we also consider a contingency game that assumes certainty one step into the future---an assumption also used in non-game-theoretic contingency planning~\cite{chen2022interactive}.
}

\smallskip\noindent
\textbf{Contingency game with oracle branching time.}
Additionally, we consider an \say{oracle} branching time estimator that recovers the true branching time by first simulating receding-horizon interaction with a nominal branching time and then extracting the time of certainty from the belief evolution in \emph{hindsight}.
Naturally, this oracle requires access to the \emph{true} human intent and hence is not realizable in practice.
Nonetheless, we include this variant to demonstrate the potential performance achievable with our interaction model. %

\subsection{Driving Scenarios}
\label{sec:driving-scenarios}
Beyond  the jaywalking example (\say{Scenario 1}) introduced in \cref{sec:contingency-features}, we evaluate our method on the following three-player scenario.

\begin{figure}[t]
    \centering
    \includegraphics[scale=\makiescale]{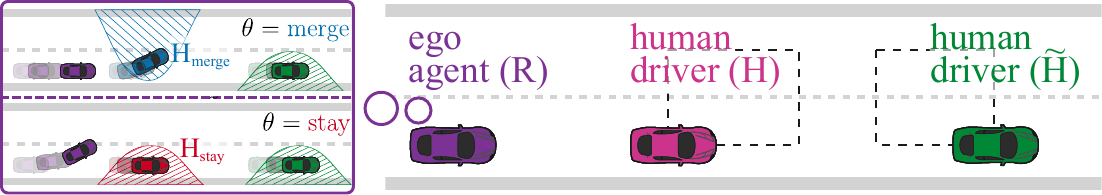}
    \caption{
    Scenario 2: an autonomous vehicle seeks to overtake slow traffic on a highway while being uncertain about the lane changing intentions of the vehicle ahead.
    }
    \label{fig:setup}
\end{figure}

\smallskip \noindent \textbf{Scenario 2: highway overtaking.} In this scenario, an autonomous vehicle attempts to overtake a human-operated vehicle with additional slow traffic in front, \cf~\cref{fig:setup}.
To perform this overtaking maneuver safely, the robot must reason about possible lane changing intentions of the other vehicle.
Since the robot is uncertain about the target lane of human-driven car in front, it maintains
a belief over the hypothesis space~$\Theta = \{\mathrm{merge}, \mathrm{stay}\}$.
We add a non-convex collision avoidance constraint between pairs of players which enforces that cars cannot be overtaken on the side of their target lane.

\smallskip \noindent \textbf{Implementation details.}
Throughout all experiments, we model cars as kinematic unicycles, and pedestrians as planar point masses.
All systems evolve in discrete-time with time discretization of~\SI{0.2}{\second} and agents plan over a horizon of~$T=25$ time steps.
\revised{In both scenarios, road users' costs comprise of a quadratic control penalty and their intent $\theta\in\Theta$ dictates a goal position for pedestrians and a reference lane for cars via a quadratic state cost.}
Collision avoidance constraints are shared between all agents.

\section{Simulated Interaction Results}
\label{sec:simulation-results}

\revised{
The following evaluations are designed to support the claims that (\textbf{C1}) contingency games close the gap between the two extremes shown in~\cref{fig:front_fig}: providing more efficient plans than fixed-uncertainty games at higher levels of safety than certainty-equivalent games; and that (\textbf{C2}) our branching time heuristic improves the performance of contingency games over a naive fixed branching time estimate of~$\branchTime = 2$.
}

\begin{figure}[t]
    \centering
    \includegraphics[scale=\largermakiescale]{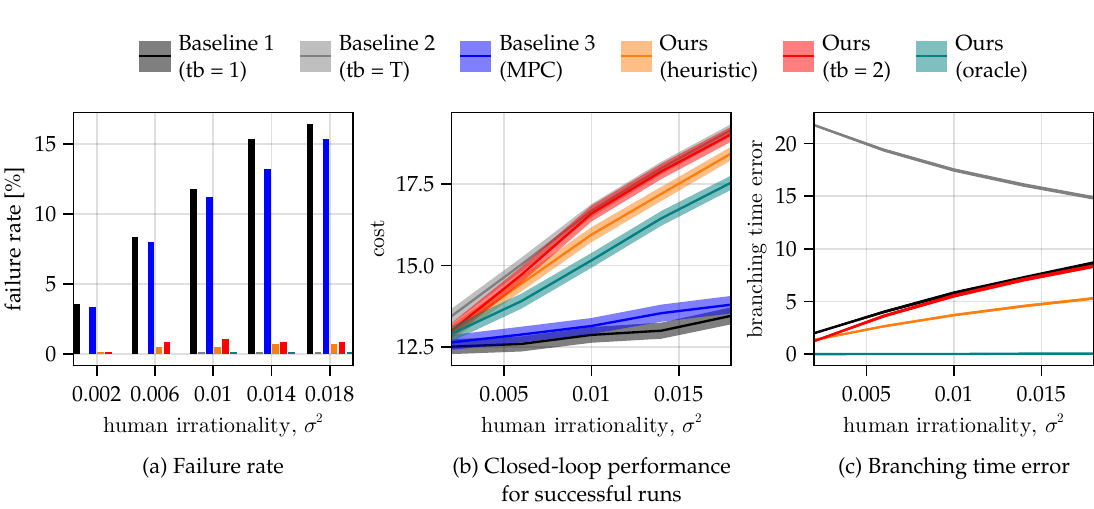}
    \caption{\revised{Quantitative closed-loop results for the jaywalking example.}}
    \label{fig:crosswalk-receding-horizon-quantitative}
\end{figure}

\begin{figure}[t]
    \centering
    \includegraphics[scale=\largermakiescale]{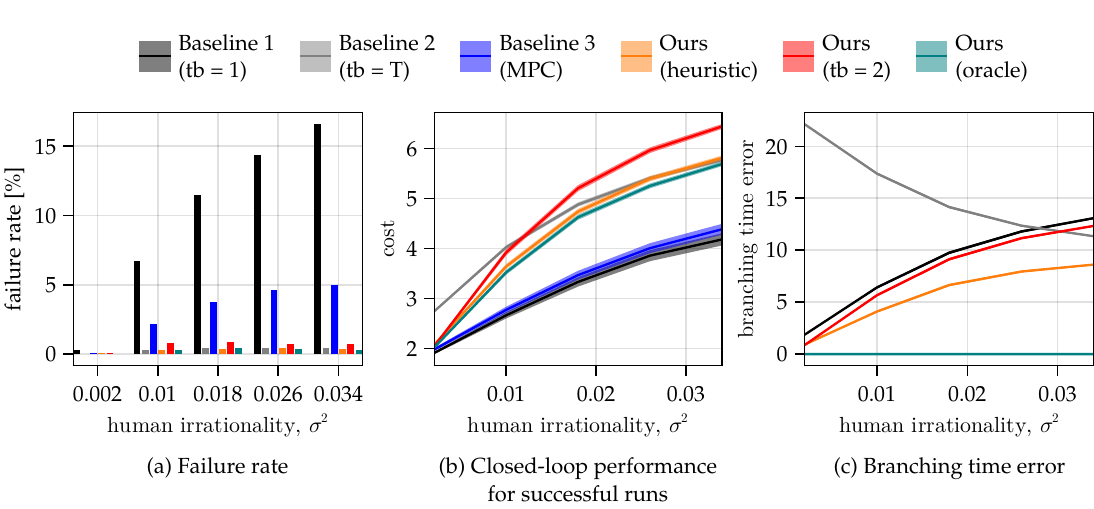}
    \caption{\revised{Quantitative closed-loop results for the overtaking example.}}
    \label{fig:overtaking-receding-horizon-quantitative}
\end{figure}

\smallskip \noindent \textbf{Data collection.}
We evaluate all methods in a large-scale Monte Carlo study as follows.
For each scenario, we simulate receding-horizon interactions of~\SI{6}{\second} duration.
As in the open-loop evaluation of \cref{sec:contingency-features}, we repeat the simulation for 70 initial states of each non-ego agent.
These initial states are drawn from a uniform grid over the state regions shown in \revised{\cref{fig:open-loop-cost-gap-spatial}a and~\ref{fig:setup}}.
We generate the ground-truth human behavior from a game solution at each fixed hypothesis $\hat{\beliefParam} \in \Theta$.
Since this true human intent is initially unknown to the robot, the robot starts with a uniform belief.
To test the methods under varying information gain dynamics---and thereby varying branching times---we consider five levels of human rationality, $\sigma^2$, parameterizing an isotropic observation model, \ie, $\Sigma = \sigma^2\identity$ where $\identity$ denotes the identity matrix.
In contrast to the open-loop evaluation of \cref{sec:contingency-features}, here the robot re-plans at every time step with the latest online estimate of the belief and branching time.
\revised{\Cref{fig:crosswalk-receding-horizon-quantitative,fig:overtaking-receding-horizon-quantitative} summarize the results of this Monte Carlo study.%
}

\smallskip \noindent
\textbf{Quantitative results.}
\revised{In terms of safety, Figures~\ref{fig:crosswalk-receding-horizon-quantitative}a and \ref{fig:overtaking-receding-horizon-quantitative}a show that the methods making a single prediction about the future, Baseline~1 and Baseline~3, fail significantly more often than the remaining approaches, all of which achieve failure rates below~1\% across all levels of human rationality.
In terms of efficiency, Figures~\ref{fig:crosswalk-receding-horizon-quantitative}b and \ref{fig:overtaking-receding-horizon-quantitative}b show that contingency games incur a lower interaction cost than the more conservative fixed-uncertainty Baseline~2.
However, this performance advantage relies on a dynamic branching time estimate, as indicated by the performance advantage of Ours~(heuristic) over Ours~($\branchTime = 2$).
Finally, Ours (oracle) further improves efficiency due to the tight branching time estimate (\cf~Figures~\ref{fig:crosswalk-receding-horizon-quantitative}c and \ref{fig:overtaking-receding-horizon-quantitative}c), demonstrating the potential of our interaction model.
In summary, these results support our claims \textbf{C1} and \textbf{C2} above.
}


\begin{figure}[t]
    \centering
    \includegraphics[scale=\makiescale]{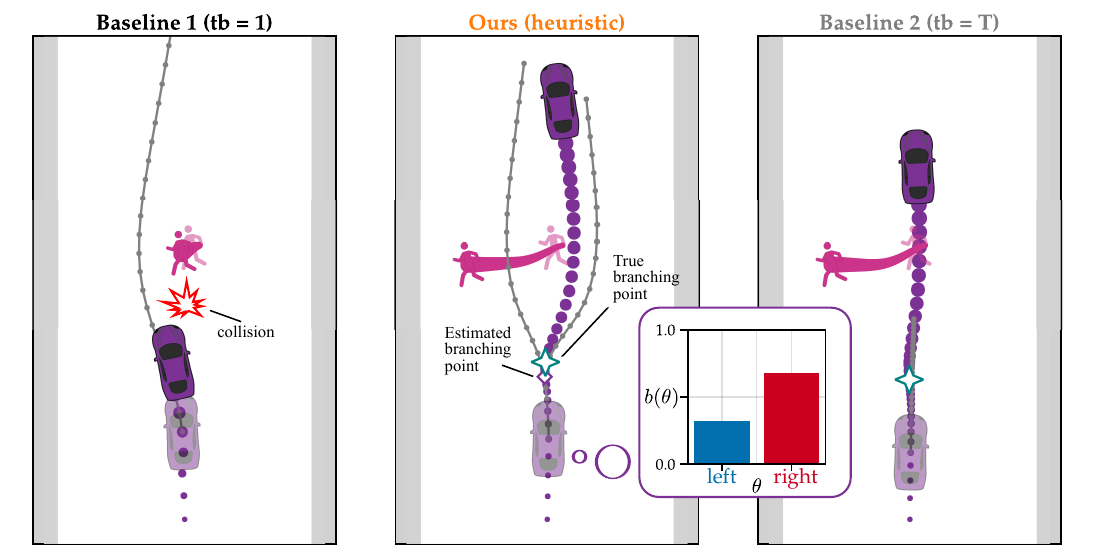}
    \caption{\revised{Qualitative closed-loop results for the jaywalking example.}}
    \label{fig:crosswalk-receding-horizon-qualitative}
\end{figure}

\begin{figure}
\centering
\includegraphics[width=\linewidth]{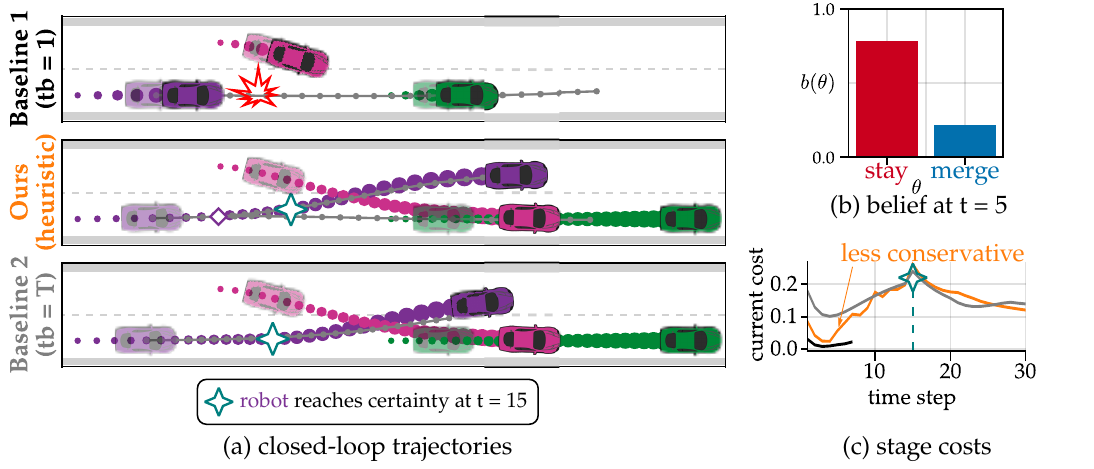}
\caption{\revised{Qualitative closed-loop results for the overtaking example.}}
\label{fig:overtaking-receding-horizon-qualitative}
\end{figure}

\smallskip \noindent \textbf{Qualitative results.}
\revised{
To further contextualize these results with respect to claim~\textbf{C1}, we visualize examples of the closed-loop behavior generated by our method, the certainty-equivalent Baseline 1, and the fixed-uncertainty Baseline 2 in~\cref{fig:crosswalk-receding-horizon-qualitative,fig:overtaking-receding-horizon-qualitative}.
}
Here, we show both the sequence of states traced out by all players and the robot's plan five time steps into the interaction.
\revised{
In both scenarios, the robot's initial observations cause its belief to favor an incorrect hypothesis.
This belief prompts Baseline~1 to commit to an unsafe strategy, causing a collision with the human.
Baseline~2, on the other hand, brakes conservatively in the face of this belief and only accelerates once the uncertainty fully resolves.
Finally, our contingency game planner \emph{anticipates} the future information gain and avoids excessive braking while remaining safe.
}

\section{Limitations \& Future Work}
Even with a simple branching time heuristic, our approach \revised{outperforms the baselines}.
Nonetheless, the observed performance gains for the branching time \say{oracle} motivate further research into more precise estimators.
\revised{
Beyond that, in this work we assumed \revisedFinal{access to a fixed set of suitable intent hypotheses and our approach relies on receding-horizon re-planning to adapt to changes of this set.
Future work should seek to automate the discovery of intent hypotheses,} test our approach in scenarios with more complex intent dynamics, and consider extensions that explicitly capture these effects in the contingency game.
}
\revised{
Furthermore, the complexity of our approach is proportional to the product of the number of individual intents of each player, and the technique we employ to solve these games generally scales cubically with regards to total strategy size.
Future work could consider employing a learning-based predictor~\cite{roh2021multimodal} to automatically identify high-likelihood intents in complex scenarios and sub-select local players~\cite{chahine2023local}.
Finally, future work may extend our approach to continuous hypothesis spaces, e.g., by sampling as in \cite{paccagnan2019scenario,le2021lucidgames}.
}

\section{Conclusion}
\label{sec:conclusion}

We present contingency games, a game-theoretic motion planning framework that enables a robot to efficiently interact with other agents in the face of uncertainty about their intents.
By capturing simplified belief dynamics, %
our method allows a robot to anticipate future changes in its belief during strategic interactions. 
In detailed simulated driving experiments, we characterized both the qualitative behavior induced by contingency games and the quantitative performance gains with respect to non-contingent \revised{baselines}.

\bibliographystyle{IEEEtranN}
{
\balance
\fontsize{9}{9}\selectfont
\bibliography{glorified,references}
}

\end{document}